\def\year{2021}\relax
\documentclass[letterpaper]{article} 
\usepackage{aaai21}  
\usepackage{times}  
\usepackage{helvet} 
\usepackage{courier}  
\usepackage[hyphens]{url}  
\usepackage{graphicx} 
\usepackage{subcaption}
\usepackage{natbib}  
\usepackage{caption} 
\usepackage{latexsym}
\usepackage{amsmath}

\newcommand{\ccirc}{\kern0.5ex\vcenter{\hbox{$\scriptstyle\circ$}}\kern0.5ex}

\usepackage{tabularx,booktabs}
\usepackage{multirow}
\usepackage{algorithm,algorithmic}
\usepackage{bbm}
\usepackage[switch]{lineno}
\newcolumntype{C}{>{\centering\arraybackslash}X}
\usepackage{enumerate}

\title{PPKE: Knowledge Representation Learning by Path-based Pre-training}

\author{Bin He\textsuperscript{\rm 1}, Di Zhou\textsuperscript{\rm 1}, Jing Xie\textsuperscript{\rm 2}\thanks{\ \ This work is done when Jing Xie is an intern at Huawei Noah's Ark Lab.}, Jinghui Xiao\textsuperscript{\rm 1}, Xin Jiang\textsuperscript{\rm 1}, Qun Liu\textsuperscript{\rm 1} \\   
}
\affiliations{
	\textsuperscript{\rm 1}Huawei Noah’s Ark Lab\\
	\textsuperscript{\rm 2}School of Computer Science, Harbin Institute of Technology\\ 
	\{hebin.nlp,zhoudi7,xiaojinghui4,jiang.xin,qun.liu\}@huawei.com\\ 
	\{xiejing.hit\}@gmail.com 
}

\begin{document}
	\maketitle
	\begin{abstract}
		Entities may have complex interactions in a knowledge graph (KG), such as multi-step relationships, which can be viewed as graph contextual information of the entities.
		Traditional knowledge representation learning (KRL) methods usually treat a single triple as a training unit, and neglect most of the graph contextual information exists in the topological structure of KGs.
		In this study, we propose a Path-based Pre-training model to learn Knowledge Embeddings, called PPKE, which aims to integrate more graph contextual information between entities into the KRL model.
		Experiments demonstrate that our model achieves state-of-the-art results on several benchmark datasets for link prediction and relation prediction tasks, indicating that our model provides a feasible way to take advantage of graph contextual information in KGs.
	\end{abstract}
	
	\section{Introduction}
	
	Knowledge graphs (KGs), such as WordNet \cite{miller1998wordnet}, Freebase \cite{bollacker2008freebase} and Wikidata \cite{vrandevcic2014wikidata}, aggregate a large amount of human knowledge and express in a structured way.
	The large number of triples in these KGs have constructed a complex knowledge network, but it is far from complete.
	In recent years, knowledge graph completion (KGC) tasks have attracted great attention.
	
	\begin{figure}[!t]
		\centering
		\includegraphics[width=0.9\linewidth]{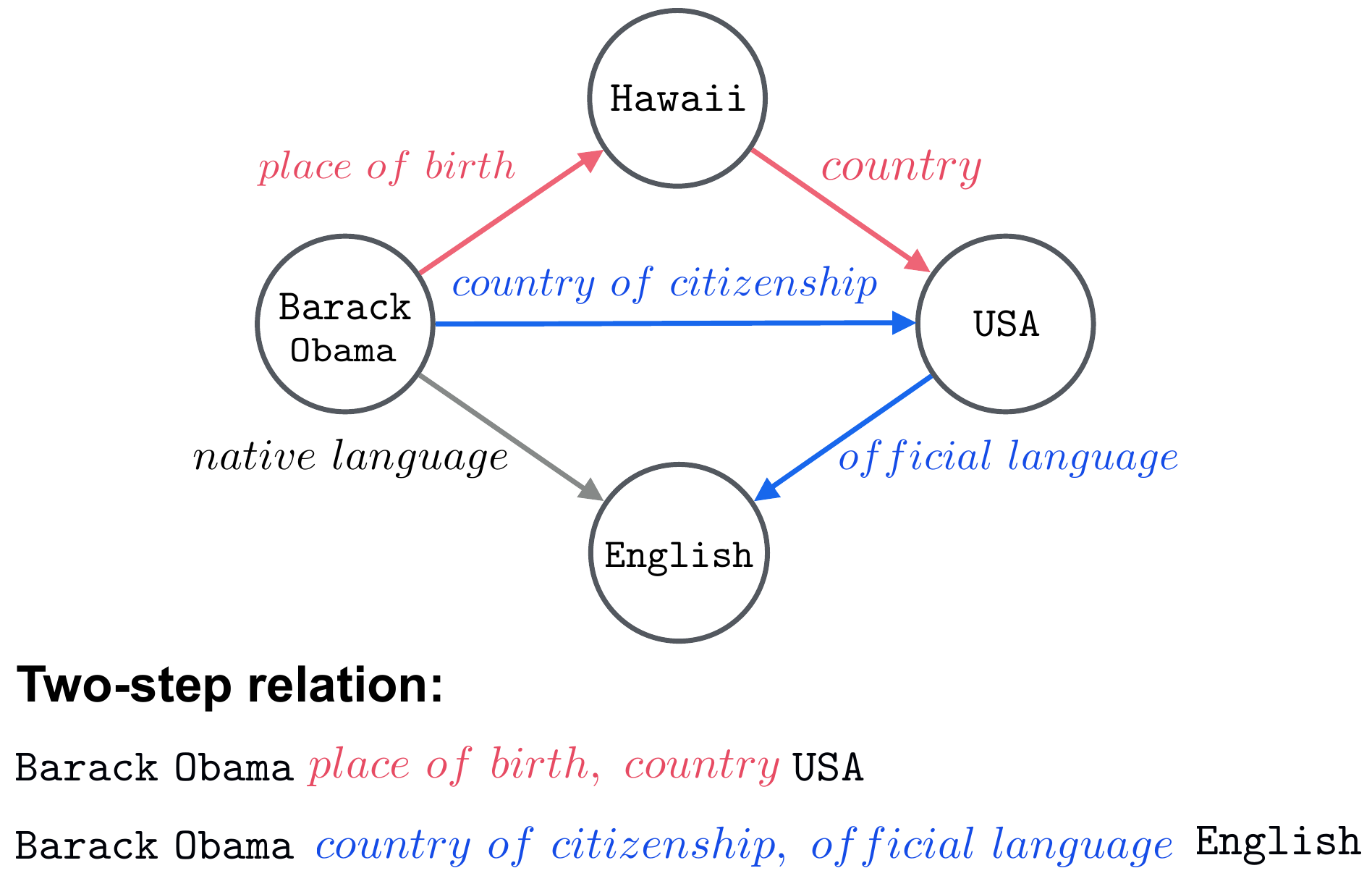}
		\caption{\label{fig:gci} An illustration of knowledge triples and relation paths. Circles are entities and arrows are relations.}
	\end{figure}
	
	Despite new state-of-the-art (SOTA) models \cite{trouillon2016complex,sun2018rotate,yao2019kgbert,wang2019coke,ji2020survey} emerge constently, most methods ignore the topological structure information of the KGs. 
	Relation paths are the most common topological structure in KGs, and Figure~\ref{fig:gci} shows some relation path instances. 
	$\scriptstyle (\mathtt{Barack\ Obama},\ country\ of\ citizenship,\ \mathtt{USA})$ is a relation triple, while $\scriptstyle \mathtt{Barack\ Obama} \xrightarrow{place\ of\ birth} \mathtt{Hawaii} \xrightarrow{country} \mathtt{USA}$ is a two-step relation path.
	Similar to word context in language models \cite{peters2018deep}, relation paths can be considered as one kind of contextual information in KGs.
	We call it ``graph contextual information''.
	And Harris's famous distributional hypothesis  \cite{harris1954distributional,firth1957synopsis} can also be extend to knowledge graphs: \textit{you shall know an entity by the relationships it involves.}
	Although these two kinds of contextual information are similar, the latter has its own specialities.
	In knowledge graphs, not all relation paths are meaningful.
	For example, $\scriptstyle \mathtt{Inception} \xrightarrow{director} \mathtt{Christopher\ Nolan} \xrightarrow{has\ child} \mathtt{Rory\ Nolan}$ is a valid relation path, but this does not indicate that there must be a relationship between $\scriptstyle\mathtt{Inception}$ and $\scriptstyle\mathtt{Rory\ Nolan}$.
	Unreliable relation paths are common in knowledge graphs, and  \citet{lin2015modeling} found that it is necessary to select reliable relation paths for knowledge representation learning.
	They learn inference patterns between relations and paths to utilize knowledge contained in relation paths.
	Instead of relying on inference patterns, we propose PPKE, a path-based pre-training approach that integrates  graph contextual information contained in relation paths into the model parameters.
	We think this is a more general way to develop the unexploited graph contextual information.
	During the path-based pre-training procedure,  two-step relation paths are extracted from the knowledge graph and fed into the pre-training module with original triples.
	Then, the pre-trained model can be finetuned for downstream KGC tasks, such as link prediction and relation prediction.
	Our contributions are as follows:
	\begin{itemize}
		\item We propose a path-based pre-training approach for knowledge representation learning, which provides a feasible way to develop more graph contextual information in KGs.
		\item Experiments on several benchmark datasets show that PPKE achieves new SOTA results on link prediction and relation prediction tasks.
	\end{itemize}

\section{Related Work}

In the literature,  relation paths have been utilized to improve the performance of KG-related tasks.
One of the representative models is PtransE \cite{lin2015modeling} , which learns inference patterns from relation paths to improve the knowledge base completion tasks.
In this work, a path-constraint resource allocation algorithm is proposed to measure the weights of inference patterns, and semantic composition of relation embeddings is utilized to represent relation paths.
Despite its success, the modeling objects are limited to the inference patterns between relations and paths, and it did not model  the contextual information that implicited in paths.

Recently, several KRL methods have attempted to introduce more contextual information into knowledge representations.
Relational Graph Convolutional Networks (R-GCNs) \cite{schlichtkrull2018modeling} is proposed to learn entity embeddings from their incoming neighbors, which greatly enhances the information interaction between related triples.
\citeauthor{nathani-etal-2019-learning} \shortcite{nathani-etal-2019-learning} further extend the information flow from 1-hop in-entities to n-hop during the learning process of entity representations.
Moreover, contextualized knowledge representation learning method (CoKE) is proposed to model the contextual nature of triples and relation paths \cite{wang2019coke}.
However, these methods do not experiment with benchmark datasets to verify that graph context information can improve the model performance \cite{sun2019re}.
We believe that the information contained in knowledge graphs has not been sufficiently exploited.
In this study, we develop a knowledge graph pre-training model to integrate more graph contextual information, and utilize this model to benefit KG-related tasks through a finetuning procedure.
	
	\section{Methodology}
	
	In this section, we will introduce our path-based pre-training model and how to finetune this model for downstream KGC tasks.
	Figure~\ref{fig:pretrain} presents the model architecture of path-based pre-training.
	
	\subsection{Path-based Pre-training}
	
	\paragraph{Input Representation}
	
	We denote relation paths as $\{h, r_1, \dots, r_n, t\}$, where $h, t$ means head and tail entity, $r_i$ is the $i^{\mathtt{th}}$-step relationship between the entity pair and $n$ is the path distance.
	When $n$ equals to 1, it represents a triple $\{h,r,t\}$.
	To distinguish the role of different entities and relations in the input sequence, position embeddings should be assigned to each input element.
	Besides, we move entities to the front of the input sequence to avoid the position bias in different sample lengths.
	Triples and relation paths are transformed into a unified format
	\begin{equation}
	\boldsymbol{x}=\{h, t, r_1, \dots, r_n\},
	\end{equation}
	and the input representation is denoted as $\mathbf{E}=\{\mathbf{E}^h, \mathbf{E}^t, \mathbf{E}^{r_1}, \dots, \mathbf{E}^{r_n}\}$.
	
	
	
	
	\paragraph{Masked Entity Prediction}
	In pre-trained language models, ``masked subword prediction'' is one of the pre-training tasks, which randomly mask subwords in a given input sequence and predict them \cite{devlin2019bert}.
	Inspired by this task, we choose entities as the masked objects.
	In our framework, only one entity (head entity or tail entity) is masked in each input sample because the input length is relatively short.
	Take the relation path $\scriptstyle (\mathtt{Barack\ Obama},\ place\ of\ birth,\ country,\ \mathtt{USA})$ as an example, we formalize this sample into $\scriptstyle \mathtt{[Barack\ Obama]\ [USA]}\ [place\ of\ birth]\ [country]$, then mask one of the entities and let the model make predictions, i.e.,
	
	$\scriptstyle \mathbf{Input}=\mathtt{[Barack\ Obama]}\ \mathtt{[MASK]}\ [place\ of\ birth]\ [country]$
	
\begin{figure}[!t]
		\centering
		\includegraphics[width=0.78\linewidth]{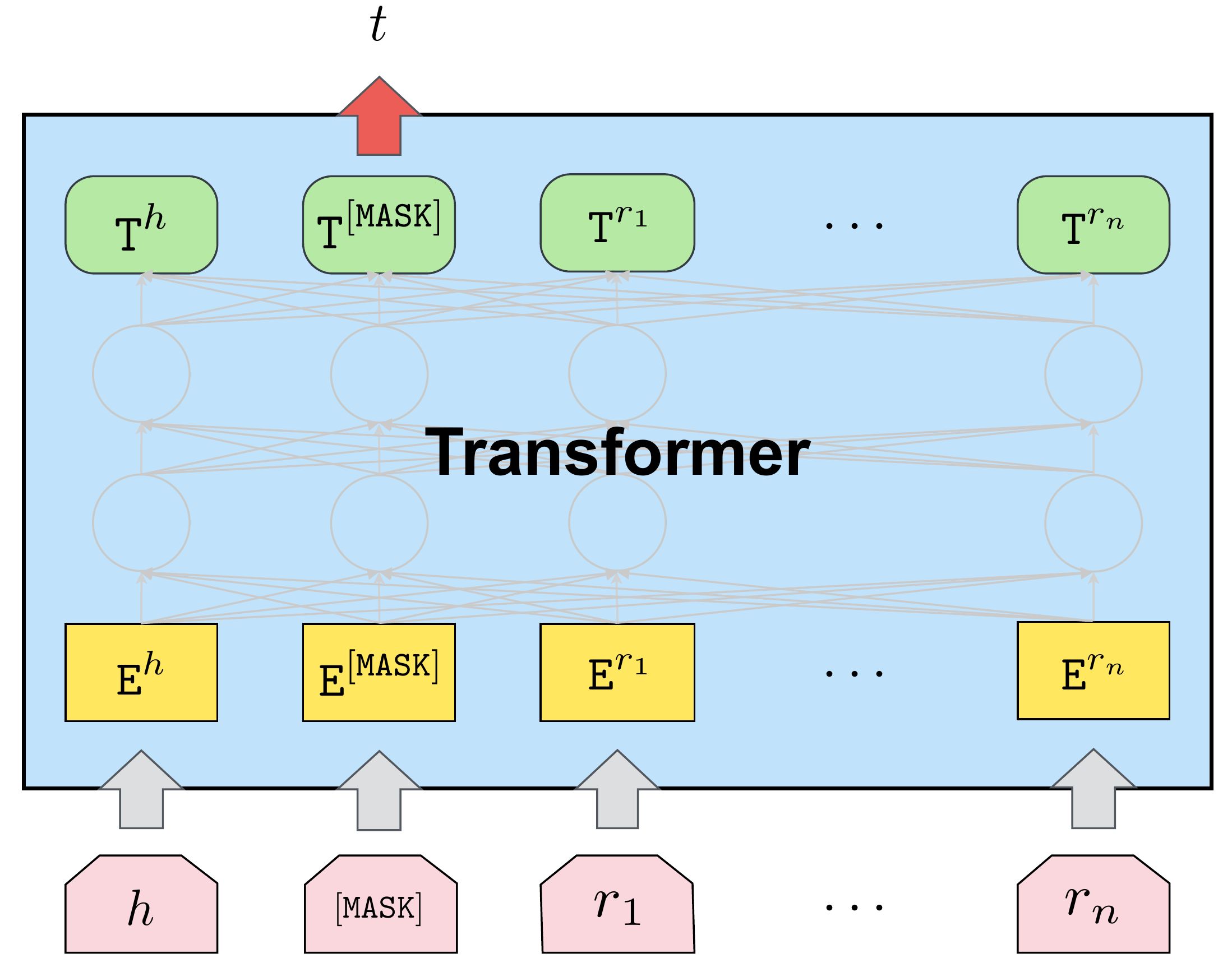}
		\caption{\label{fig:pretrain} The model architecture of path-based pre-training. $h, t$ denote the head and tail entity, $r_1,\dots, r_n$ is the $n$-step relationship between the entity pair.}
	\end{figure}

	$\scriptstyle \mathbf{Label}=\mathtt{[USA]}$
	
	\noindent where $[\cdot]$ denotes an input element.
	As shown in Figure~\ref{fig:pretrain}, we use a Transformer encoder \cite{vaswani2017attention} to learn the graph contextual information of the inputs.
	Assuming $e$ is the masked entity, and $\mathbf{T}^{ \mathtt{[MASK]}}$is the output hidden state of input $\scriptstyle \mathtt{[MASK]}$, the prediction objective is to calculate the prediction probability of entity $e$ and maximize it.
	This process can be formalized as follows:
	\begin{align}
	\boldsymbol{p}^\mathtt{[MASK]} &= \mathrm{softmax}(\mathbf{T}^{\mathtt{[MASK]}}\cdot\mathbf{V}^{\mathcal{E}}),\\
	\mathcal{L}&=-\log \boldsymbol{p}^\mathtt{[MASK]}_e ,
	\end{align}
	where $\boldsymbol{p}^\mathtt{[MASK]}$ is the probability vector of all candidate entities in the entity vocabulary, and $\mathbf{V}^{\mathcal{E}}$ is the entity embedding matrix.
	We utilize the maximum likelihood estimation (MLE) method to learn the model parameters:
	\begin{equation}
	\hat{\boldsymbol{\theta}}=\mathop{\mathrm{argmax}}_{\boldsymbol{\theta}}\sum_{(\boldsymbol{x},\boldsymbol{m}_e)\in \mathcal{X}}\log p(\boldsymbol{x}_e|\boldsymbol{x}\ccirc \boldsymbol{m}_e;\boldsymbol{\theta}),
	\end{equation}
	where $\boldsymbol{m}_e$ is the masking vector, and $\boldsymbol{x}_e$ represents the target entity.
	
	\begin{table*}
	\centering
	\begin{tabular}{llcr}
		\hline
		\bf Dataset  & \bf \# Ent / \# Rel & \bf \# Train / Valid / Test & \bf \# Quad \\
		\hline
		FB15k-237 & 14,541 / 237 & 272,115 / 17,535 / 20,466 & 17,765,062\\
		WN18RR & 40,943 / 11 & 86,835 / 3,034 / 3,134 & 236,475\\
		FB15k & 14,951 / 1,345 & 483,142 / 50,000 / 59,071 & 81,916,109\\
		\hline
	\end{tabular}
	\caption{\label{tab:dataset} \small Statistics of the datasets. ``Quad'' means quadruple, i.e., 2-step relation path between two entities, such as $(h, r_1, r_2, t)$. Datasets can be downloaded from this repository\footnote{\url{https://github.com/thunlp/OpenKE}}.}
\end{table*}

\begin{table*}
	\centering
	\begin{tabular}{lllllllll}
		\hline
		\bf Method & \multicolumn{4}{c}{\bf FB15k-237} & \multicolumn{4}{c}{\bf WN18RR}   \\
		\cmidrule(lr){2-5} \cmidrule(lr){6-9}
		& MRR & H@1 &  H@3 &  H@10 & MRR & H@1 &  H@3 &  H@10 \\ 
		\hline
		TransE~\cite{bordes2013translating}  & 29.0 & 19.9 & - & 47.1& 46.6 & 42.3 & - & 55.5 \\
		ConvE~\cite{dettmers2018convolutional}    & 31.6 & 23.9 & 35.0 & 49.1& 46.0 & 39.0 & 43.0 & 48.0 \\
		ConvKB~\cite{nguyen2018novel}   & \bf 39.6 & - & - & 51.7& 24.8 & - & - & 52.5  \\
		R-GCN~\cite{schlichtkrull2018modeling}   & 24.9 & 15.1 & 26.4 & 41.7 & - & - & - & - \\
		RotatE~\cite{sun2018rotate}  & 33.8 & 24.1 & 37.5 & 53.3& 47.6 & 42.8 & 49.2 & 57.1 \\
		CoKE~\cite{wang2019coke}  & 36.1 & 26.9 & 39.8 & 54.7& 47.5 & 43.7 & 49.0 & 55.2 \\
		\hline
		CoKE (ours) & 35.9 & 26.6 & 39.6 & 54.5& 47.5 & 43.6 & 48.8 & 55.4  \\
		PPKE  & 36.4 & \bf 27.0 & \bf 40.0 & \bf 55.1& \bf 48.5 & \bf 43.9 & \bf 50.2 & \bf 57.3 \\
		\hline
	\end{tabular}
	\caption{\label{tab:lp_new}  \small Link prediction results on WN18RR and FB15k-237. The results of TransE  are reported by OpenKE \cite{han2018openke}. The original CoKE model was implemented using deep learning framwork PaddlePaddle, and we re-implement this model using TensorFlow. The scores of remaining baselines follow their original papers.}
\end{table*}
	
	\subsection{Finetuning}
	
	\paragraph{Link Prediction}
	This task is to predict the tail entity given head entity and relation ($(h, r, ?) \rightarrow t$), or to predict the head entity given relation and tail entity($(?, r, t) \rightarrow h $) . 
	Aligning with the pre-trained model, the task objective is to predict the masked head(tail) entity.
	An example of the inputs is as follows:
	
	$\scriptstyle \mathbf{Input}_\mathrm{head}=\mathtt{[MASK]}\ \mathtt{[USA]}\ [country\ of\ citizenship]$
	
	$\scriptstyle\mathbf{Label}_\mathrm{head}=\mathtt{[Barack\ Obama]}$
	
	$\scriptstyle\mathbf{Input}_\mathrm{tail}=\mathtt{[Barack\ Obama]}\ \mathtt{[MASK]}\ [country\ of\ citizenship]$
	
	$\scriptstyle\mathbf{Label}_\mathrm{tail}=\mathtt{[USA]}$
	
	\noindent Given training dataset $\mathcal{D}_{\boldsymbol{xe}}$, the training objective is to maxinize the log-likelihood of the masked entity conditioned on the masked triple:
	\begin{equation}
	\hat{\boldsymbol{\theta}}_{\boldsymbol{x}\rightarrow\boldsymbol{e}}=\mathop{\mathrm{argmax}}_{\boldsymbol{\theta}_{\boldsymbol{x}\rightarrow\boldsymbol{e}}}\sum_{\boldsymbol{x}\in \mathcal{D}_{\boldsymbol{xe}}}\log p(e|\boldsymbol{x};\boldsymbol{\theta}_{\boldsymbol{x}\rightarrow\boldsymbol{e}}).
	\end{equation}
	
	\paragraph{Relation Prediction}
	
	This task is to predict the relation given head entity and tail entity ($(h, ?, t)\rightarrow r $).
	In this task, the task objective is to predict the masked relation, and an example of the inputs is as follows:
	
	$\scriptstyle\mathbf{Input}_\mathrm{rel}=\mathtt{[Barack\ Obama]}\ \mathtt{[USA]}\ \mathtt{[MASK]}$
	
	$\scriptstyle\mathbf{Label}_\mathrm{rel}=[country\ of\ citizenship]$
	
	\noindent Given training dataset $\mathcal{D}_{\boldsymbol{xr}}$, the training objective is to maxinize the log-likelihood of the masked relation conditioned on the masked triple:
	\begin{equation}
	\hat{\boldsymbol{\theta}}_{\boldsymbol{x}\rightarrow\boldsymbol{r}}=\mathop{\mathrm{argmax}}_{\boldsymbol{\theta}_{\boldsymbol{x}\rightarrow\boldsymbol{r}}}\sum_{\boldsymbol{x}\in \mathcal{D}_{\boldsymbol{xr}}}\log p(r|\boldsymbol{x};\boldsymbol{\theta}_{\boldsymbol{x}\rightarrow\boldsymbol{r}}).
	\end{equation}
	
	\section{Experiments}
	
	\subsection{Dataset}
	
	We conduct link prediction experiments on two widely-used benchmark datasets: FB15k-237 \cite{toutanova2015observed}, which is a subset of Freebase; WN18RR \cite{dettmers2018convolutional}, which is a subset of WordNet.
	We also conduct relation prediction on FB15k \cite{bordes2013translating} and FB15k-237.
	The detailed statistics of these datasets are listed in Table~\ref{tab:dataset}. 
	
	\subsection{Settings}
	
	During the model pre-training, only two-step relation paths (or quadruples) are extracted from KGs.
	We count the number of quadruples in different training sets and the statistics is shown in Table~\ref{tab:dataset}.
	For WN18RR, all quadruples are involved in pre-training.
	For FB15k-237 and FB15k, due to limitations of computing resources, 2 and 5 million quadruples are randomly chosen, respectively.
	In the pre-training phase, the maximum sequence length is set to 4.
	We use a batch size of 512 for WN18RR, and 4096 for FB15k-237 and FB15k.
	The pre-training epoch number is set to 800.
	In this study, the Transformer layer number, head number and hidden size are set to 6, 4, 256, which are the same as that used in \cite{wang2019coke}.
	%
	%
	
	%
	%
	
	\subsection{Results}

	
	
	Table~\ref{tab:lp_new} shows the model comparison on two link prediction datasets.
	We list six previous state-of-the-art models as baselines to compare the link prediction performance with our model, and the results demonstrate that PPKE outperforms on most evaluation metrics.
	Compared with CoKE, PPKE  achieves significant improvement on WN18RR, which improves 1.0\% and 2.1\% on MRR and Hits@10, respectively.
	Besides, in FB15k-237, PPKE outperforms CoKE by 0.3\% and 0.4\% on MRR and Hits@10.
	Table~\ref{tab:rp} presents  relation prediction results on FB15k and FB15k-237.
	As we can see, KG-BERT and CoKE outperform PTransE and ProjE on FB15k, and PPKE achieves better results on both FB15k and FB15k-237.
	These experimental results verify that our model has the ability to utilize graph contextual information to improve the performance of link prediction and relation prediction.
	
	
		\begin{table*}
		\centering
		\begin{tabular}{lcc}
			\hline
			\bf Method & \bf FB15k & \bf FB15k-237  \\
			\hline
			PTransE (ADD, 2-step) \cite{lin2015modeling}  & 93.6  & - \\
			PTransE (ADD, 3-step) \cite{lin2015modeling}   & 94.0  & - \\
			ProjE (pointwise) \cite{shi2017proje}  & 95.6 & -  \\
			ProjE (listwise) \cite{shi2017proje}  & 95.7 & - \\
			ProjE (wlistwise) \cite{shi2017proje}  & 95.6 & - \\
			KG-BERT (b) \cite{yao2019kgbert}  & 96.0 & -  \\
			\hline
			CoKE (ours)  & 96.6 & 95.5 \\
			PPKE  & \bf 96.8 & \bf 95.7 \\
			\hline
		\end{tabular}
		\caption{\label{tab:rp}  \small Relation prediction results (Hits@1) on FB15k and FB15k-237. The baselines are taken from original papers.}
	\end{table*}

\begin{table*}
	\centering
	\begin{tabular}{lcccc}
		\hline
		\bf Method & \multicolumn{2}{c}{\bf FB15k-237} & \multicolumn{2}{c}{\bf WN18RR}  \\
		\cmidrule(l){2-3} \cmidrule(l){4-5}
		& MRR &  Hits@10 & MRR &  Hits@10\\ 
		\hline
		CoKE (ours) & 35.9 & 54.5  & 47.5 & 55.4 \\
		PPKE (w/o pretraining) & 35.7 & 54.2 & 47.1 & 55.0 \\
		PPKE (w/ pretraining)  & \bf 36.4 & \bf 55.1 & \bf 48.5 & \bf 57.3 \\
		\hline
	\end{tabular}
	\caption{\label{tab:kgt_ablation} Ablation study on the link prediction task.}
\end{table*}
	
	\subsection{Discussion}
	
	\paragraph{Does the path-based pre-training work?}
	
	Table~\ref{tab:kgt_ablation} shows the ablation study on whether to use path-based pre-training in PPKE.
	Without pre-training, our model achieves similiar results of CoKE.
	The slightly lower scores may be caused by hyperparameter searching, because we follow the hyperparameters used in CoKE.
	After using pre-training, the scores of Hits@10 in WN18RR and FB15k-237 improve by 2.3\% and 0.9\%, respectively.

	\paragraph{What the model learned during path-based pre-training?}
	
\begin{figure}[!t]
		\centering
		\includegraphics[width=0.75\linewidth]{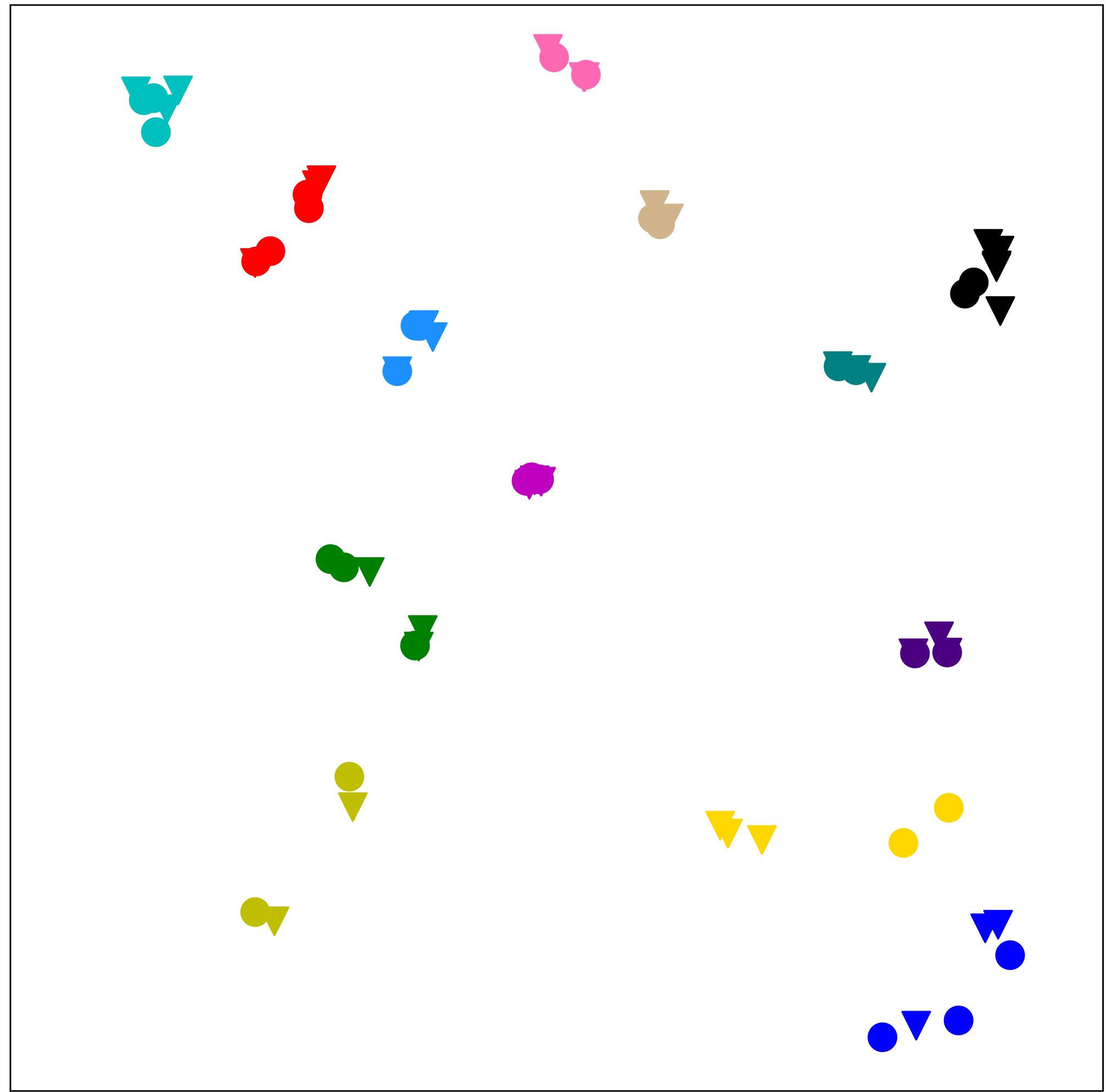}
		\caption{\label{fig:path_emb} Visualization for masked tail entities. Different colors mean different sample groups. Samples of the same color share the same head and tail entity.  {\it Round point} represents hidden state representation for the masked tail entity in a triple while {\it inverted triangle} is for that in a quadruple.}
	\end{figure}

	We collect some groups of triples and quadruples from the pre-training pool.
	In each sample group, triples and quadruples share the same head and tail entity.
	After masking the tail entity for each sample group, the samples are fed into the pre-trained model to generate the output hidden state of the masked entity.
	Figure~\ref{fig:path_emb} illustrates these hidden state representations via t-SNE \cite{maaten2008visualizing}.
	In most sample groups, the cohesion degree is high, indicating our model has integrated two-step graph contextual information into the model parameters.
	
	\section{Conclusion and Future Work}
	
	We propose a novel approach to integrate graph contextual information into a path-based pre-training model, focusing on modeling one-step and two-step relations between entities.
	Experiments show our model outperforms previous state-of-the-art methods,
	which validates the intuition that graph contextual information is beneficial to knowledge graph completion tasks.
	
	In the follow-up work, we will try to add relation prediction objective into the pre-training procedure, and larger quantity or wider variety of graph contextual information will be explored.
	
	
	\section*{Acknowledgments}
	
	Thank Chao Xing for his insightful suggestions for this work.
	
	\bibliographystyle{aaai21}
	\bibliography{PPKE}
	
\end{document}